\documentclass{article} 
\usepackage{iclr2026_conference,times}
\usepackage{amsthm}
\usepackage{newtxmath}

\usepackage{amsmath,amsfonts,bm}

\def\eqref#1{equation~\ref{#1}}

\def\1{\bm{1}}

\def\va{{\bm{a}}}

\def\vg{{\bm{g}}}
\def\vh{{\bm{h}}}

\def\vp{{\bm{p}}}

\def\vr{{\bm{r}}}

\def\vw{{\bm{w}}}
\def\vx{{\bm{x}}}
\def\vy{{\bm{y}}}
\def\vz{{\bm{z}}}

\def\mA{{\bm{A}}}

\def\mH{{\bm{H}}}
\def\mI{{\bm{I}}}
\def\mJ{{\bm{J}}}

\def\mP{{\bm{P}}}

\def\mW{{\bm{W}}}
\def\mX{{\bm{X}}}

\DeclareMathAlphabet{\mathsfit}{\encodingdefault}{\sfdefault}{m}{sl}
\SetMathAlphabet{\mathsfit}{bold}{\encodingdefault}{\sfdefault}{bx}{n}

\def\gF{{\mathcal{F}}}

\newcommand{\R}{\mathbb{R}}

\usepackage{hyperref}
\usepackage{bookmark}
\usepackage{url}
\usepackage{booktabs}
\usepackage{graphicx}
\usepackage{float}
\usepackage{subcaption}
\usepackage{xcolor}
\usepackage{xspace}
\usepackage{tikz}
\usetikzlibrary{arrows.meta,positioning}
\usepackage{amsmath,mathtools}
\usepackage{tabularx}
\usepackage{array}
\usepackage{multirow}
\usepackage{makecell}

\newcommand{\mhc}{\textit{m}HC\xspace}
\newcommand{\Hpre}{\mH^{\text{pre}}}
\newcommand{\Hpost}{\mH^{\text{post}}}
\newcommand{\Hres}{\mH^{\text{res}}}
\newcommand{\neff}{n_{\mathrm{eff}}}

\title{How Does {\textit{m}HC} Use Its Residual Streams? Selective Routing and Near-Identity Mixing}

\author{
Pengxiang Zhao$^{1}$ \quad Xing Li$^{1}$ \quad Xianzhi Yu$^{1}$ \quad Wei Guo$^{1}$ \quad Zhenhua Dong$^{1}$\\
$^{1}$Huawei Technologies Co., Ltd. \\
\texttt{\{zhao.pengxiang, li.xing2\}@huawei.com}
}

\iclrfinalcopy 
\begin{document}

\maketitle

\begin{abstract}
Hyper-Connections and their manifold-constrained variant (\mhc) widen a residual pathway from one stream to \(n\), yet how trained models use this capacity remains unclear: how broadly blocks read and write, how strongly the residual pathway mixes streams, and whether the streams carry distinct representations.
We examine these properties in the four-stream residual pathway of DeepSeek-V4-Flash using effective stream counts, cross-stream residual weights, and inter-stream cosine similarity.
Read/write routing is concentrated but varies across depth: a typical attention or FFN site effectively uses about two streams, while the dominant stream changes across layers and the representations remain directionally distinct.
Residual mixing is modest and occurs primarily in early layers; in layers 22--42, the pathway mostly carries each stream forward separately.
Targeted interventions establish the functional significance of these patterns.
Replacing the late mixers by identity increases C4 perplexity by only $1.9\%$ and preserves the six-task average score, whereas replacing the early mixers increases perplexity by $41\%$.
Fixing each early mixer to its C4 diagnostic mean increases perplexity by only $0.2\%$ and reduces the average score by $0.25$ percentage points, showing that its site-specific structure matters more than its token-wise variation on the evaluated metrics.
Likewise, retaining the three largest routing weights per token at every site increases perplexity by at most $2.7\%$ and changes the average score by at most $0.4$ points.
Thus, the studied model realizes only part of the flexibility afforded by four-stream \mhc: individual blocks rarely require all four streams, and late residual mixing provides little measured benefit.
\end{abstract}

\begin{figure}[H]
    \centering
    \includegraphics[width=\linewidth]{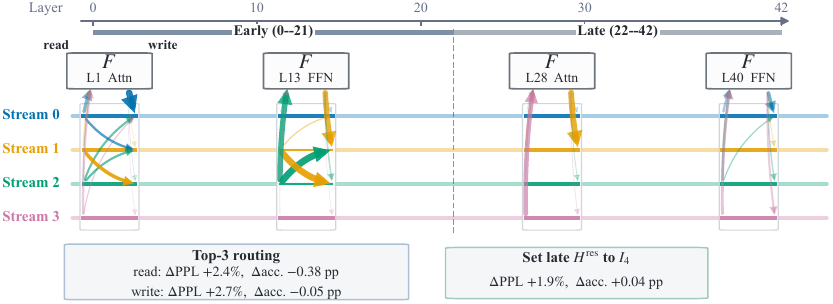}
    \captionsetup{font=small}
    \caption{
    \textbf{Realized multi-stream computation in DeepSeek-V4-Flash.}
    At four sites spanning depth and both sublayer types, color identifies the source stream, arrows show the direction of information flow, and line width encodes the measured token-averaged routing or residual weight.
    Routing concentrates on changing stream subsets, whereas residual exchange largely disappears after mid-depth; the callouts give the exact intervention effects on C4 perplexity and the six-task average score.
    }
    \label{fig:overview}
\end{figure}

\section{Introduction}
\label{sec:intro}

Hyper-Connections (HC)~\citep{zhu2024hyperconnections} generalize the standard single-stream residual pathway~\citep{he2016resnet} to $n$ coupled streams.
At each sublayer, a read map aggregates the streams into a single branch input, a write map distributes the branch output back across the streams, and a residual mixer linearly recombines the incoming stream states along the skip pathway.
Manifold-constrained Hyper-Connections (\mhc)~\citep{xie2026mhc} are designed to improve training stability by constraining each residual mixer to be doubly stochastic, thereby limiting signal amplification as these mixers compose across depth.
This design widens the state propagated across depth while retaining a single computation branch at each sublayer.

However, \mhc makes broad routing and cross-stream mixing possible, while training determines how and where the resulting model uses this flexibility.
Because the read, write, and residual maps are token-conditioned and learned at each layer, their realized behavior can vary across both tokens and depth.
Accordingly, the read and write maps may distribute branch computation broadly or concentrate it on a subset of streams; the residual pathway may mix the stream states or propagate them separately; and the streams may encode distinct or redundant representations.

Recent work on 120M- and 360M-parameter HC models built with nanoGPT~\citep{karpathy2022nanogpt} finds a dominant-stream regime: read/write signals and interpretable features concentrate in one stream, while residual mixing remains close to identity~\citep{alimaskina2026streamcollapse}.
Because model scale and training regime may shape how the learned maps allocate computation across streams, it remains unclear whether this regime is characteristic of trained \mhc models or only one of several possible organizations of multi-stream computation.
Moreover, these structural patterns do not by themselves establish post-training functional redundancy.
Determining this requires testing whether weak routing weights can be pruned and whether near-identity residual mixers can be replaced by exact identity maps without degrading model quality.

We address \textit{both gaps} with a true-forward analysis of DeepSeek-V4-Flash, a 284B-parameter \mhc model pretrained on more than 32T tokens~\citep{deepseekai2026deepseekv4}.
Our framework measures effective read/write width, cross-stream residual weights, and inter-stream representation similarity across tokens, sublayers, and depth.
These depth-resolved diagnostics distinguish a single globally dominant stream from shifting stream use across layers and determine whether the streams retain distinct representations.
We pair them with targeted inference-time interventions to connect the observed routing and mixing patterns to model quality.

Applying this framework reveals concentrated but depth-varying routing: a typical attention or FFN site effectively uses about two of the four streams, while the dominant stream changes across layers and the stream representations remain directionally distinct.
Routing ablations show that this concentration creates local redundancy but does not reduce the pathway to two streams.
Retaining the three largest realized read or write weights for every token at every site increases perplexity on C4~\citep{raffel2020t5} by at most $2.7\%$ and changes the average downstream score by at most $0.4$ percentage points; retaining only two weights per token instead increases perplexity by $12$--$14\%$ and reduces the average downstream score by $2.9$--$6.6$ percentage points.

Residual mixing also varies with depth.
Cross-stream residual weights are modest overall and most pronounced in layers $0$--$21$ of the 43-layer model, whereas the residual mixers in layers $22$--$42$ remain close to the identity.
Residual-mixer ablations confirm that this depth profile has functional significance.
Replacing the mixers in layers $22$--$42$ by exact identity maps increases C4 perplexity by only $1.9\%$ and preserves the average score across six downstream benchmarks, whereas replacing those in layers $0$--$21$ increases perplexity by $41\%$.
Yet fixing each early mixer to its C4 diagnostic mean raises perplexity by only $0.2\%$ and reduces the six-task average score by $0.25$ points, indicating that its learned site-specific structure accounts for most of its measured benefit, while token-wise variation contributes little on the evaluated metrics.
Figure~\ref{fig:overview} provides a data-driven overview of this analysis, tracing realized routing and residual exchange across representative depths and linking both to their measured intervention effects.

In summary, our contributions are threefold:
\begin{itemize}
    \item \textbf{True-forward analysis of multi-stream computation.}
    We introduce a framework that resolves read/write routing, residual mixing, and stream representations across tokens, sublayers, and depth, and connects these measurements to model quality through inference-time interventions.

    \item \textbf{A depth-resolved characterization of DeepSeek-V4-Flash.}
    We show that its four streams exhibit locally concentrated but depth-varying routing, retain distinct representations, and exchange information primarily in the first half of the network.

    \item \textbf{Functional localization of routing and mixing capacity.}
    Controlled interventions identify where additional routing weights and learned residual mixing affect model quality, revealing local redundancy in read/write maps and limited functional dependence on late-layer residual mixing.
\end{itemize}

\section{Background and Related Work}
\label{sec:background}

\subsection{Residual Streams and Hyper-Connections}
\begin{figure}[t]
    \centering
    \resizebox{\columnwidth}{!}{\begin{tikzpicture}[
    x=1cm,
    y=1cm,
    font=\sffamily,
    >=Latex,
    flow/.style={draw=black!72, line width=0.8pt,
                 -{Latex[length=1.65mm,width=1.1mm]},
                 shorten <=0.7pt, shorten >=1.2pt},
    stream/.style={draw=black!55, fill=black!3, rounded corners=1.2pt,
                   minimum width=1.02cm, minimum height=0.34cm,
                   inner sep=1.2pt, font=\scriptsize},
    map/.style={rounded corners=2.5pt, minimum width=1.10cm,
                minimum height=0.72cm, align=center, inner sep=2pt,
                line width=0.8pt, font=\scriptsize},
    state/.style={rounded corners=2pt, minimum width=0.92cm,
                  minimum height=0.34cm, align=center, inner sep=1.3pt,
                  font=\scriptsize},
    note/.style={font=\fontsize{6.4}{7.2}\selectfont, text=black!62},
]

\node[stream] (xin1) at (0, 0.98) {$\vx_l^{(1)}$};
\node[font=\scriptsize, text=black!60] (xindots) at (0, 0.62) {$\vdots$};
\node[stream] (xinn) at (0, 0.26) {$\vx_l^{(n)}$};
\node[font=\fontsize{6.7}{7.4}\selectfont\bfseries, text=black!68] at (0, 1.36) {incoming streams};

\node[stream] (xout1) at (8.75, 0.98) {$\vx_{l+1}^{(1)}$};
\node[font=\scriptsize, text=black!60] (xoutdots) at (8.75, 0.62) {$\vdots$};
\node[stream] (xoutn) at (8.75, 0.26) {$\vx_{l+1}^{(n)}$};
\node[font=\fontsize{6.7}{7.4}\selectfont\bfseries, text=black!68] at (8.75, 1.36) {outgoing streams};

\coordinate (xin) at (0.62,0.62);
\coordinate (xout) at (8.22,0.62);

\node[map, draw=blue!72!black, fill=blue!8] (read) at (1.80,1.65)
    {$\Hpre_l$\\[-1pt]{\color{black!62}\tiny $1\!\times\! n$}};
\node[note, text=blue!60!black] at (1.80,2.16) {READ};
\node[state, draw=blue!55!black, fill=blue!3] (branchin) at (3.15,1.65)
    {$\Hpre_l\mX_l$};
\node[map, draw=black!60, fill=black!5, minimum width=1.18cm] (branch) at (4.70,1.65)
    {$\gF(\cdot,\mW_l)$};
\node[map, draw=orange!82!black, fill=orange!12] (write) at (6.30,1.65)
    {$(\Hpost_l)^\top$\\[-1pt]{\color{black!62}\tiny $n\!\times\!1$}};
\node[note, text=orange!72!black] at (6.30,2.16) {WRITE};

\node[map, draw=teal!72!black, fill=teal!9, minimum width=1.28cm] (res) at (4.28,-0.05)
    {$\Hres_l$\\[-1pt]{\color{black!62}\tiny $n\!\times\! n$}};
\node[note, text=teal!65!black] at (4.28,-0.56) {RESIDUAL MIXER};

\node[circle, draw=black!65, fill=white, minimum size=0.46cm,
      inner sep=0pt, line width=0.8pt, font=\small] (sum) at (7.75,0.62) {$+$};

\draw[flow] (xin) -- ++(0.28,0) |- (read.west);
\draw[flow] (read.east) -- (branchin.west);
\draw[flow] (branchin.east) -- (branch.west);
\draw[flow] (branch.east) -- (write.west);
\draw[flow] (write.east) -- ++(0.62,0)
    node[note, above, midway] {$n\!\times\! C$} -| (sum.north);

\draw[flow] (xin) -- ++(0.28,0) |- (res.west);
\draw[flow] (res.east) -- ++(2.05,0)
    node[note, above, midway] {$n\!\times\! C$} -| (sum.south);
\draw[flow] (sum.east) -- (xout);

\end{tikzpicture}}
    \caption{Data flow through HC residual sublayer $l$. The read map $\Hpre_l$ forms the computation-branch input, the write map $\Hpost_l$ distributes the branch output across streams, and the residual mixer $\Hres_l$ carries forward and linearly recombines the incoming stream states.}
    \label{fig:hc-schematic}
\end{figure}
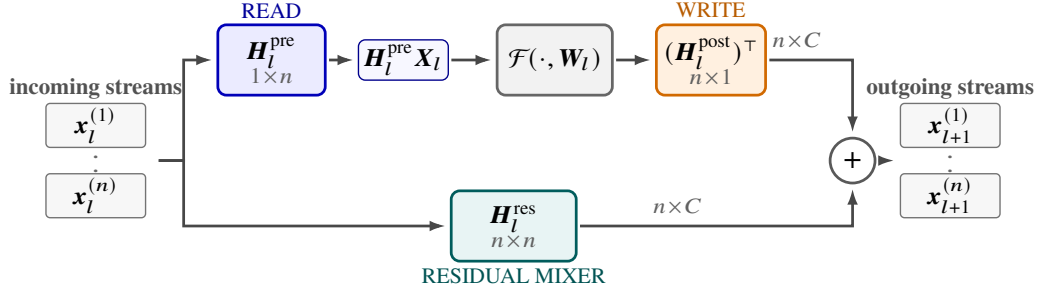

Consider a residual sublayer with hidden dimension $C$.
A standard residual architecture propagates a single state, whereas Hyper-Connections (HC) maintain $n$ coupled states $\mX_l \in \R^{n \times C}$ at sublayer $l$~\citep{zhu2024hyperconnections}.
Here and below, $l$ indexes residual sublayers, including the attention and FFN sublayers within each Transformer layer; we suppress the token index for clarity.
Each sublayer applies three maps to this state: a read map $\Hpre_l \in \R^{1 \times n}$, a write map $\Hpost_l \in \R^{1 \times n}$, and a residual mixer $\Hres_l \in \R^{n \times n}$.
In HC, each map combines learned static parameters with a token-dependent component.
We refer to the resulting value for a given token and sublayer as the \emph{realized} map.
Figure~\ref{fig:hc-schematic} summarizes the resulting data flow.

Let $\gF(\cdot,\mW_l)$ denote the computation performed by the sublayer.
Its multi-stream update is
\begin{equation}
\label{equ:hc-update}
    \mX_{l+1}
    =
    \Hres_l\mX_{l}
    +
    (\Hpost_l)^\top\gF(\Hpre_l\mX_{l}, \mW_l),
\end{equation}
where $\Hpre_l\mX_l$ forms the input to the computation branch, $(\Hpost_l)^\top\gF(\Hpre_l\mX_l,\mW_l)$ distributes its output across the streams, and $\Hres_l\mX_l$ carries forward and linearly recombines the incoming stream states.

\subsection{Manifold-Constrained Hyper-Connections}
Across successive residual sublayers, the residual mixers compose multiplicatively along the skip pathway.
For sublayers $l,\ldots,L-1$, the resulting cumulative mixer is
\begin{equation}
\label{equ:cumulative}
    \Pi_{L:l}^{\mathrm{res}}
    \coloneqq
    \mH_{L-1}^{\mathrm{res}}
    \cdots
    \mH_l^{\mathrm{res}}.
\end{equation}
Without constraints on the individual mixers, the norm of this product can grow rapidly with depth, amplifying signals propagated along the residual pathway.
To control this cumulative transformation, \mhc constrains every realized residual mixer to the Birkhoff polytope~\citep{xie2026mhc}:
\begin{equation}
\label{equ:birkhoff}
    \mathcal{B}_n
    \coloneqq
    \left\{
    \mH\in\R^{n\times n}
    \;\middle|\;
    \mH\mathbf{1}_n=\mathbf{1}_n,\;
    \mathbf{1}_n^\top\mH=\mathbf{1}_n^\top,\;
    \mH\ge 0
    \right\},
\end{equation}
where $\mathbf{1}_n$ is the all-ones vector.
In practice, \mhc implements this projection using Sinkhorn--Knopp normalization~\citep{sinkhorn1967concerning,xie2026mhc}.
Under the exact constraint, the Birkhoff polytope is closed under multiplication, and every $\mH\in\mathcal{B}_n$ has spectral norm one.
Consequently, the cumulative mixer $\Pi_{L:l}^{\mathrm{res}}$ remains non-expansive, preventing signal amplification along the skip pathway.
Although $\mI_n$ is a valid doubly stochastic mixer, \mhc learns $\Hres_l$ rather than fixing it to $\mI_n$ to enable cross-stream information exchange~\citep{xie2026mhc}.

\subsection{Analysis of Learned Stream Utilization}
The original HC and \mhc studies establish the optimization and performance properties of multi-stream residual pathways~\citep{zhu2024hyperconnections,xie2026mhc}.
Recent analysis of $120$M- and $360$M-parameter HC models trained with nanoGPT further identifies a dominant-stream regime in which read/write signals and interpretable features concentrate in one stream while the residual mixers remain close to identity~\citep{karpathy2022nanogpt,alimaskina2026streamcollapse}.
Breaking initialization symmetry can mitigate this concentration~\citep{alimaskina2026streamcollapse}.
These findings establish stream collapse as one possible organization of learned HC computation.
Our study examines how routing, residual mixing, and stream representations are organized in a separately trained large-scale \mhc model and tests the functional significance of these patterns through inference-time interventions.

\section{Measuring Realized Multi-Stream Computation}
\label{sec:framework}

For each token at residual sublayer \(l\), the realized maps
\(\Hpre_l\), \(\Hpost_l\), and \(\Hres_l\) determine how branch computation is routed and how residual states are mixed across streams.
We characterize this computation along three axes: the breadth of read/write routing, the amount of cross-stream residual mixing, and the similarity between stream representations.

\subsection{Read/Write Routing Breadth}
We characterize read/write routing at each sublayer from two complementary perspectives: whether tokens share the same dominant stream and how broadly the routing weights are distributed across streams.
For diagnostic sequence $m$, let $h^q_{l,m,t,j}$ denote the realized routing weight assigned to stream $j$ for its $t$-th token at sublayer $l$, where $q\in\{\textit{pre},\textit{post}\}$ and $j\in\{1,\ldots,n\}$.
The realized read and write weights are nonnegative under the \mhc parameterization, so we normalize them to obtain the token-level load distribution
\begin{equation}
\label{equ:routing-normalization}
    p^q_{l,m,t,j}
    \coloneqq
    \frac{h^q_{l,m,t,j}}
         {\sum_{k=1}^{n}h^q_{l,m,t,k}}.
\end{equation}
We summarize the resulting token-level distributions with two sublayer-level statistics.

\paragraph{Winner consistency.}
For each token, we identify the stream receiving the largest routing weight.
For an input sequence $m$ containing $T_m$ tokens, its winner consistency is the largest fraction of tokens sharing the same winner:
\begin{equation}
\label{equ:winner-consistency}
    c^q_{l,m}
    \coloneqq
    \max_{j\in\{1,\ldots,n\}}
    \frac{1}{T_m}
    \sum_{t=1}^{T_m}
    \mathbb{I}
    \left[
        j=\arg\max_k p^q_{l,m,t,k}
    \right],
\end{equation}
where $c^q_{l,m}\in[1/n,1]$.
We report $c^q_l=\frac{1}{M}\sum_{m=1}^{M}c^q_{l,m}$ over the $M$ diagnostic sequences.
A value of $1$ means that all tokens within every sequence share the same winner, while lower values indicate that the dominant stream varies across tokens.

\paragraph{Load-effective stream count.}
We first aggregate the normalized routing weights within each sequence:
\begin{equation}
\label{equ:stream-load}
    \bar p^q_{l,m,j}
    \coloneqq
    \frac{1}{T_m}\sum_{t=1}^{T_m}p^q_{l,m,t,j}.
\end{equation}
The effective number of streams carrying this load is
\begin{equation}
\label{equ:neff}
    \neff(\bar{\vp}^q_{l,m})
    \coloneqq
    \frac{1}{\sum_{j=1}^{n}(\bar p^q_{l,m,j})^2}
    \in[1,n].
\end{equation}
The effective stream count equals \(1\) when all routing load falls on one stream and \(n\) when the load is uniform.
As with winner consistency, we report this quantity averaged over the $M$ diagnostic sequences.

\subsection{Cross-Stream Residual Mixing}

We quantify how strongly each residual sublayer mixes its incoming streams by measuring the deviation of its realized residual mixer from the identity.
Let $\Hres_{l,t}\in\R^{n\times n}$ denote the realized residual mixer for token $t$ at sublayer $l$.
We define its identity deviation as
\begin{equation}
\label{equ:delta}
    \delta(\Hres_{l,t})
    \coloneqq
    \frac{1}{n^2}
    \sum_{i=1}^{n}\sum_{j=1}^{n}
    \left|
        (\Hres_{l,t})_{ij}
        -
        \mathbb{I}[i=j]
    \right|.
\end{equation}
The metric is zero when each stream is carried forward independently through the identity mixer.
Under the exact doubly stochastic constraint, it is proportional to the total off-diagonal weight and therefore increases with cross-stream residual mixing.
We average identity deviation across tokens within each sublayer and examine its profile across depth.

\subsection{Inter-Stream Representation Similarity}

We assess whether the residual streams carry distinct representations by measuring their pairwise alignment.
Let $\vx^{(s)}_{l,t}\in\R^C$ denote the hidden state carried by stream $s$ for token $t$ at sublayer $l$.
We define the mean pairwise cosine similarity as
\begin{equation}
\label{equ:stream-cosine}
    s_{l,t}
    \coloneqq
    \frac{2}{n(n-1)}
    \sum_{1\le i<j\le n}
    \frac{
        \langle \vx^{(i)}_{l,t}, \vx^{(j)}_{l,t} \rangle
    }{
        \|\vx^{(i)}_{l,t}\|_2
        \|\vx^{(j)}_{l,t}\|_2
    }.
\end{equation}
Higher values indicate more closely aligned stream representations, whereas lower values indicate greater directional separation.
We average this similarity across tokens within each sublayer and examine how it evolves across depth.

\subsection{Inference-Time Interventions}
\label{sec:interventions}

The diagnostics above describe how the trained model uses its residual streams, but establishing which components are functionally necessary requires a counterfactual test: would model quality change if selected routing paths or cross-stream residual interactions were suppressed?
We perform this test through two targeted inference-time interventions without retraining: routing sparsification and residual-mixer replacement.

\begin{figure}[t]
    \centering
    \resizebox{0.84\linewidth}{!}{\begin{tikzpicture}[
    x=1cm, y=1cm, font=\sffamily, >=Latex,
    flow/.style={draw=black!52, line width=0.68pt,
                 -{Latex[length=1.45mm,width=0.95mm]}},
    weight/.style={draw=blue!66!black, fill=blue!10,
                   minimum width=0.48cm, minimum height=0.34cm,
                   inner sep=0pt, rounded corners=1.1pt,
                   font=\rmfamily\fontsize{5.15}{5.7}\selectfont},
    output/.style={draw=teal!65!black, fill=teal!9},
    removed/.style={draw=black!18, fill=black!3, text=black!32},
    oplabel/.style={font=\fontsize{5.2}{5.8}\selectfont,
                    text=orange!68!black},
]

\node[weight] at (0.28,0.84) {0.41};
\node[weight] at (0.84,0.84) {0.29};
\node[weight] at (1.40,0.84) {0.21};
\node[weight] at (1.96,0.84) {0.09};
\draw[flow] (2.34,0.84) -- (4.22,0.84);
\node[oplabel] at (3.28,1.08)
    {$\displaystyle k=3,\;\alpha=.91^{-1}$};
\node[weight,output]  at (4.54,0.84) {0.45};
\node[weight,output]  at (5.10,0.84) {0.32};
\node[weight,output]  at (5.66,0.84) {0.23};
\node[weight,removed] at (6.22,0.84) {0};

\node[weight] at (0.28,0.24) {0.41};
\node[weight] at (0.84,0.24) {0.29};
\node[weight] at (1.40,0.24) {0.21};
\node[weight] at (1.96,0.24) {0.09};
\draw[flow] (2.34,0.24) -- (4.22,0.24);
\node[oplabel] at (3.28,0.48)
    {$\displaystyle k=2,\;\alpha=.70^{-1}$};
\node[weight,output]  at (4.54,0.24) {0.59};
\node[weight,output]  at (5.10,0.24) {0.41};
\node[weight,removed] at (5.66,0.24) {0};
\node[weight,removed] at (6.22,0.24) {0};

\end{tikzpicture}}
    \caption{Scale-preserving top-$k$ routing for $k\in\{2,3\}$. Retained weights are rescaled by $\alpha$ to preserve their original sum.}
    \label{fig:topk-intervention}
\end{figure}
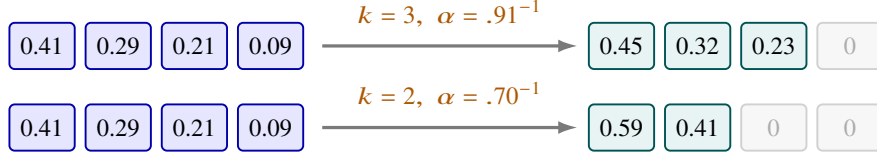

For routing sparsification, we suppress the sequence index $m$ and let $\mathcal{T}_{l,t}^{q}(k)$ index the $k$ largest realized weights of the read or write map for token $t$ at sublayer $l$.
We define the scale-preserving top-$k$ map as
\begin{align}
\alpha_{l,t}^{q}(k)
&\coloneqq
\frac{\sum_{r=1}^{n} h^{q}_{l,t,r}}
     {\sum_{r\in\mathcal{T}_{l,t}^{q}(k)} h^{q}_{l,t,r}},
\\
\widetilde h^{q}_{l,t,j}
&\coloneqq
\alpha_{l,t}^{q}(k)\,
h^{q}_{l,t,j}\,
\mathbb{I}\!\left[j\in\mathcal{T}_{l,t}^{q}(k)\right].
\label{equ:topk-intervention}
\end{align}
This intervention changes the routing support while preserving the total routing weight.
Figure~\ref{fig:topk-intervention} illustrates the operation for $k\in\{2,3\}$.
We apply it to the read and write maps separately.

For residual-mixer replacement, we substitute $\mI_n$ for $\Hres_{l,t}$ at every token within a selected depth range.
This removes learned cross-stream exchange within that range while leaving the per-stream skip paths unchanged.
We evaluate both interventions using perplexity and downstream-task scores.

\section{Experiments}
\label{sec:experiments}
We apply the framework from Section~\ref{sec:framework} to DeepSeek-V4-Flash.
We first describe the model, data, and true-forward collection protocol.
We then examine read/write routing, stream representations, and residual mixing across depth, followed by inference-time interventions on the learned routing and residual maps.

\subsection{Experimental Setup}
\label{sec:experimental-setup}

\paragraph{Model.}
We study the publicly released 0731 checkpoint of DeepSeek-V4-Flash~\citep{deepseekai2026deepseekv4}, a 43-layer Mixture-of-Experts (MoE) language model with 284B total and 13B activated parameters, pretrained on more than 32T tokens.
Its 4096-dimensional backbone combines hybrid attention and MoE feed-forward networks with four-stream \mhc residual pathways.
We instrument the attention and MoE-FFN residual sublayers in every layer, yielding 86 observation sites.

\paragraph{Data and evaluation.}
We use C4~\citep{raffel2020t5} following the sampling and evaluation protocols of GPTQ~\citep{frantar2023gptq}.
For the true-forward diagnostics, we randomly sample 512 sequences of 1,024 tokens from the first English C4 training shard.
For perplexity, we follow the GPTQ evaluation protocol~\citep{frantar2023gptq}, evaluating 256 contiguous sequences of 2,048 tokens constructed from the first English C4 validation shard.
We additionally evaluate six downstream benchmarks in the zero-shot setting: ARC-Easy and ARC-Challenge~\citep{clark2018think}, PIQA~\citep{bisk2020piqa}, HellaSwag~\citep{zellers2019hellaswag}, MMLU~\citep{hendrycks2021measuring}, and GSM8K~\citep{cobbe2021training}.
We report normalized accuracy for ARC-Easy, ARC-Challenge, PIQA, and HellaSwag, accuracy for MMLU, and exact-match accuracy for GSM8K, together with the unweighted average of the six task scores.

\paragraph{Implementation.}
We build our forward-pass instrumentation and inference-time interventions on PyTorch~\citep{paszke2019pytorch} and Hugging Face Transformers~\citep{wolf2020transformers}.
Downstream evaluation uses the LM Evaluation Harness~\citep{gao2023lmeval}.

\subsection{Organization of Read/Write Routing}
\label{sec:routing-results}

\begin{figure}[t]
    \centering
    \begin{subfigure}[t]{0.45\linewidth}
        \centering
        \includegraphics[width=\linewidth]{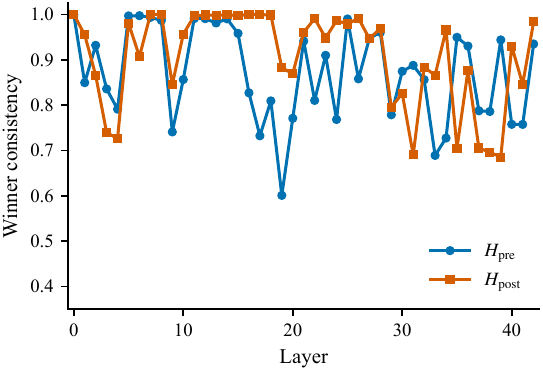}
        \caption{Winner consistency.}
        \label{fig:routing-winner}
    \end{subfigure}\hfill
    \begin{subfigure}[t]{0.45\linewidth}
        \centering
        \includegraphics[width=\linewidth]{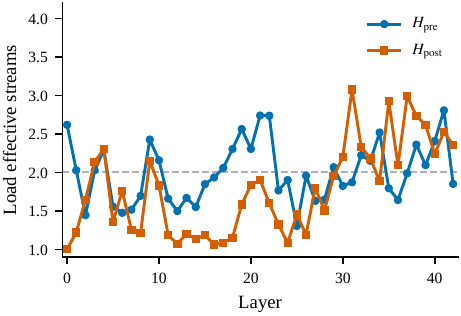}
        \caption{Load-effective stream count.}
        \label{fig:routing-effective}
    \end{subfigure}
    \caption{
    Depth profiles of routing concentration.
    Each point averages the attention and FFN sites within a layer.
    (a) Winner consistency measures the fraction of tokens sharing the most frequent dominant stream; its read/write averages are $0.871$ and $0.905$.
    (b) The load-effective stream count averages $1.998$ and $1.775$, respectively, indicating that routing typically spans only about two of the four streams.
    The dashed line in (b) marks two effective streams.
    }
    \label{fig:routing-concentration}
\end{figure}

We first characterize how the trained model routes branch computation across its four residual streams.
Figure~\ref{fig:routing-concentration} shows that routing is both narrow and largely stable across tokens at a given sublayer.
Winner consistency averages $0.871$ for the read maps and $0.905$ for the write maps, with medians of $0.931$ and $0.988$ across sublayers; thus, most maps assign a large majority of tokens within an input sequence to the same dominant stream.
The corresponding load-effective stream counts average $1.998$ and $1.775$, with medians of $1.876$ and $1.542$.
A typical sublayer therefore concentrates most of its routing load on roughly two streams, with stronger concentration in the write maps.

\begin{figure}[t]
    \centering
    \begin{subfigure}[t]{0.235\linewidth}
        \centering
        \includegraphics[width=\linewidth]{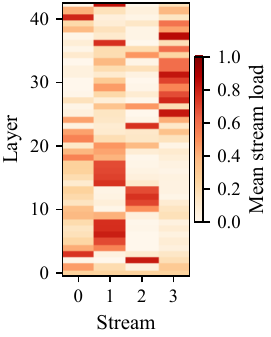}
        \caption{$\Hpre$ at attention sites.}
        \label{fig:routing-load-pre-attn}
    \end{subfigure}\hfill
    \begin{subfigure}[t]{0.235\linewidth}
        \centering
        \includegraphics[width=\linewidth]{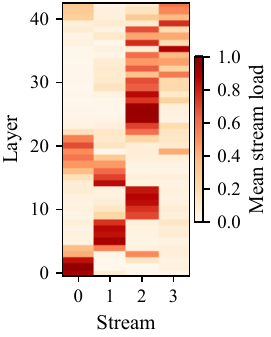}
        \caption{$\Hpre$ at FFN sites.}
        \label{fig:routing-load-pre-ffn}
    \end{subfigure}\hfill
    \begin{subfigure}[t]{0.235\linewidth}
        \centering
        \includegraphics[width=\linewidth]{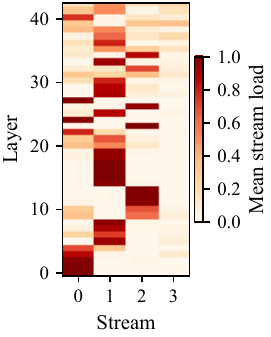}
        \caption{$\Hpost$ at attention sites.}
        \label{fig:routing-load-post-attn}
    \end{subfigure}\hfill
    \begin{subfigure}[t]{0.235\linewidth}
        \centering
        \includegraphics[width=\linewidth]{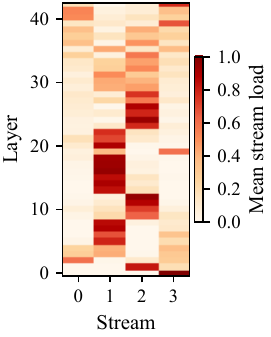}
        \caption{$\Hpost$ at FFN sites.}
        \label{fig:routing-load-post-ffn}
    \end{subfigure}
    \caption{
    Mean normalized routing load by layer and stream, shown separately for the read and write maps at attention and FFN sites.
    High-load regions form contiguous depth intervals, but their stream assignments change across layers, sublayer types, and read/write maps.
    }
    \label{fig:routing-load}
\end{figure}

This local concentration does not reduce to a single stream dominating throughout the network.
Figure~\ref{fig:routing-load} instead reveals depth-structured assignments: for a fixed map and sublayer type, $62.5\%$ of adjacent layer pairs retain the same winner, producing contiguous intervals whose dominant stream changes at several depth boundaries.
The reorganization is especially clear in the read maps.
Streams 0 and 1 dominate 32 of the 44 sites in layers 0--21, whereas streams 2 and 3 dominate 36 of the 42 sites in layers 22--42.
Every stream therefore becomes dominant in read or write routing somewhere in the network, but over different depth ranges.

Routing breadth also changes with depth, primarily through the write maps.
Their mean effective stream count increases from $1.467$ in layers 0--21 to $2.098$ in layers 22--42, while winner consistency decreases from $0.940$ to $0.870$; read width remains nearly unchanged at $1.971$ and $2.025$, respectively.
The read and write schedules are coupled but not identical: their winners differ at $38.4\%$ of the 86 sites, so a sublayer need not write primarily to the stream from which it reads.
Taken together, routing is locally narrow and largely token-stable, yet its allocation is reorganized across depth rather than collapsing onto one globally dominant stream.
Appendix~\ref{app:diagnostics-depth-site} reports the routing statistics by depth range and sublayer type, while Appendix~\ref{app:lowest-load-route} identifies the lowest-load stream at each site.

\subsection{Inter-Stream Representation Geometry}
\label{sec:representation-results}

\begin{figure}[t]
    \centering
    \begin{subfigure}[t]{0.45\linewidth}
        \centering
        \includegraphics[width=\linewidth]{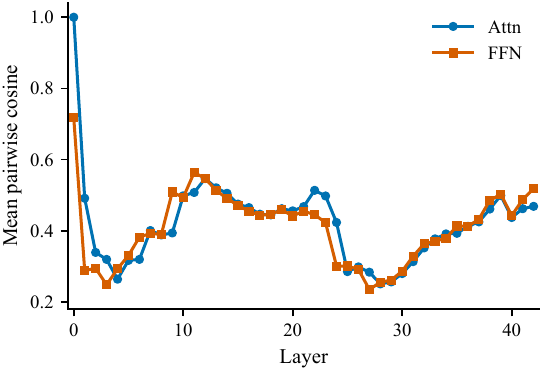}
        \caption{Mean over the six stream pairs.}
        \label{fig:stream-cosine-depth}
    \end{subfigure}\hfill
    \begin{subfigure}[t]{0.45\linewidth}
        \centering
        \includegraphics[width=\linewidth]{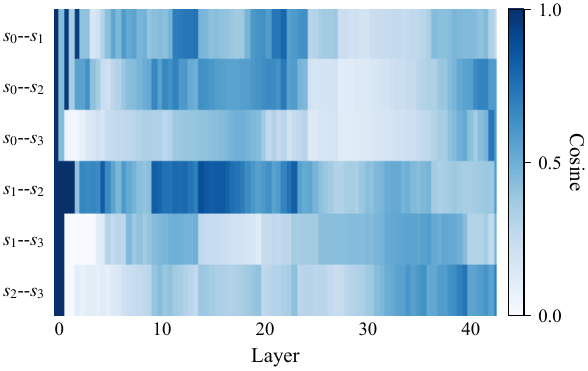}
        \caption{Pair-resolved similarity.}
        \label{fig:stream-cosine-pairs}
    \end{subfigure}
    \caption{
    Inter-stream representation similarity across depth.
    (a) Pairwise cosine similarity, averaged over tokens and the six stream pairs at each attention and FFN site.
    (b) Token-averaged cosine similarity for each stream pair, with attention and FFN sites shown in forward-pass order within each layer.
    The streams begin from identical states but rapidly diverge, and no stream pair remains uniformly aligned throughout the network.
    }
    \label{fig:stream-cosine}
\end{figure}

Having established that routing is locally concentrated, we ask whether this concentration is accompanied by redundancy in the stream representations.
Figure~\ref{fig:stream-cosine} tracks their pairwise cosine similarity across depth.
The streams are identical at the first attention site, where their mean pairwise cosine is $1.0$, but separate rapidly as the first layers transform and redistribute their states.
From layer 1 onward, mean similarity ranges from $0.235$ to $0.564$ across attention and FFN sites, with an average of $0.404$.
Their directional separation is not monotonic: the streams partially realign around layers 10--13, separate more strongly around layers 27--29, and become progressively more aligned toward the output, without returning to their initial agreement.

The pair-resolved view further shows that this geometry is heterogeneous rather than symmetric.
Beyond layer 0, pair-specific averages range from $0.288$ for streams 0 and 3 to $0.560$ for streams 1 and 2, and no pair remains uniformly aligned across depth.
Thus, local routing concentration does not coincide with representation collapse: even weakly routed streams can carry directionally distinct states. However, representational distinctness does not establish functional importance; Section~\ref{sec:intervention-results} tests whether these weak routing paths materially affect model quality.

\subsection{Depth Structure of Residual Mixing}
\label{sec:residual-results}
\begin{figure}[t]
    \centering
    \begin{subfigure}[t]{0.40\linewidth}
        \centering
        \includegraphics[width=\linewidth]{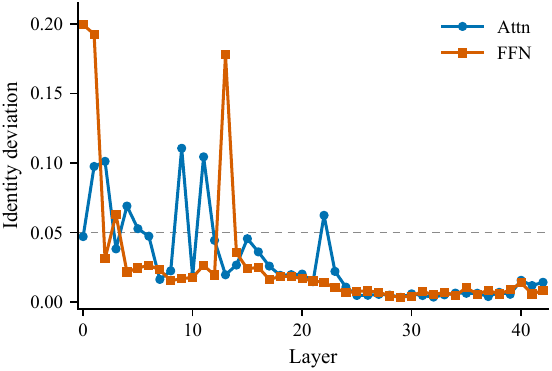}
        \caption{Identity deviation across depth.}
        \label{fig:hres-depth}
    \end{subfigure}\hfill
    \begin{subfigure}[t]{0.275\linewidth}
        \centering
        \includegraphics[width=\linewidth]{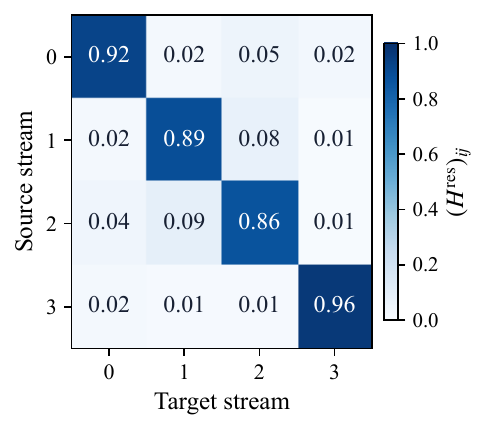}
        \caption{Layers 0--21; $\delta=0.046$.}
        \label{fig:hres-early}
    \end{subfigure}\hfill
    \begin{subfigure}[t]{0.275\linewidth}
        \centering
        \includegraphics[width=\linewidth]{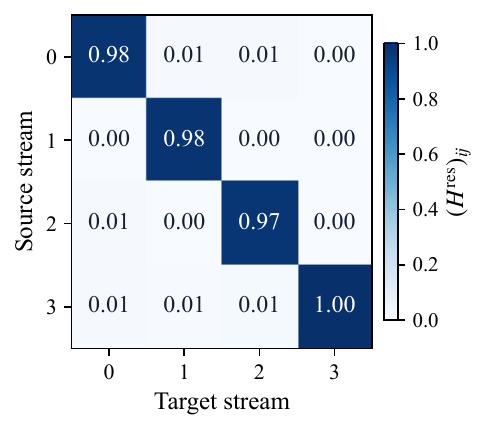}
        \caption{Layers 22--42; $\delta=0.009$.}
        \label{fig:hres-late}
    \end{subfigure}
    \caption{
    Depth structure of realized residual mixing.
    (a) Identity deviation at each attention and FFN site, averaged over diagnostic tokens.
    (b--c) Residual mixers averaged over tokens and sites in the indicated depth ranges.
    Off-diagonal transfer is localized primarily to early layers, while the later mean mixer is nearly diagonal and performs little direct cross-stream exchange.
    }
    \label{fig:hres-structure}
\end{figure}

We next examine how the strength of learned residual mixing varies across depth. Figure~\ref{fig:hres-depth} reveals pronounced variation across the network: identity deviation averages \(0.028\) across all 86 sites, but the largest deviations occur at only a small number of early sites.
After a final attention-side spike at layer 22, the mixers settle close to identity: 32 of the 40 sites in layers 23--42 have identity deviation below $0.01$.

Figures~\ref{fig:hres-early} and~\ref{fig:hres-late} show how this depthwise reduction appears in the residual mixers themselves. Over layers 0--21, the mean mixer retains visible off-diagonal weights, with an identity deviation of \(0.046\). Over layers 22--42, these weights largely vanish and the deviation falls to \(0.009\), leaving a near-diagonal map that carries each stream forward with little direct residual exchange. The complete layerwise matrices for attention and FFN sites are provided in Appendix~\ref{app:layerwise-hres} (Figures~\ref{fig:hres-attn-all} and~\ref{fig:hres-ffn-all}).

The late mixers therefore realize little of their available cross-stream flexibility despite being learned and token-dependent by design.
Their near-static identity structure motivates replacing them with \(\mI_4\) at inference, thereby avoiding the corresponding Sinkhorn--Knopp iterations; Section~\ref{sec:intervention-results} evaluates the resulting effect on model quality.

\subsection{Functional Effects of Routing and Mixing Interventions}
\label{sec:intervention-results}

The preceding diagnostics characterize realized stream use, but they do not determine which routing and mixing effects are necessary for model quality.
We therefore apply the inference-time interventions from Section~\ref{sec:interventions}, without retraining, and measure their effects on C4 perplexity and downstream-task scores in Tables~\ref{tab:intervention-ppl} and~\ref{tab:intervention-downstream}.

\begin{table}[H]
    \centering
    \caption{
    C4 perplexity under the inference-time interventions.
    ``Token mean'' replaces each selected mixer by its average over the C4 diagnostic sequences.
    }
    \label{tab:intervention-ppl}

    \footnotesize
    \renewcommand{\arraystretch}{0.8}
    \setlength{\tabcolsep}{12pt}

    \resizebox{0.8\columnwidth}{!}{%
    \begin{tabular}{
        @{\hspace{6pt}}
        lllccc
        @{\hspace{6pt}}
    }
        \toprule
        \textbf{Type}
        & \textbf{Action}
        & \textbf{Scope}
        & \textbf{PPL $\downarrow$}
        & \textbf{$\Delta$ PPL}
        & \textbf{Relative $\Delta$} \\
        \midrule

        \multicolumn{3}{@{\hspace{6pt}}l}{Baseline}
        & 11.1587
        & ---
        & --- \\

        \midrule

        \multirow{4}{*}{%
            \makecell[l]{Routing}%
        }
        & \multirow{2}{*}{top-3}
        & $\Hpre$
        & 11.4220
        & +0.2633
        & +2.4\% \\

        &
        & $\Hpost$
        & 11.4587
        & +0.3000
        & +2.7\% \\

        \cmidrule(lr){2-6}

        &
        \multirow{2}{*}{top-2}
        & $\Hpre$
        & 12.5052
        & +1.3465
        & +12.1\% \\

        &
        & $\Hpost$
        & 12.7206
        & +1.5619
        & +14.0\% \\

        \midrule

        \multirow{3}{*}{%
            \makecell[l]{Mixer}%
        }
        & \multirow{2}{*}{$\Hres \rightarrow \mI_4$}
        & 22--42
        & 11.3728
        & +0.2141
        & +1.9\% \\

        &
        & 0--21
        & 15.7835
        & +4.6248
        & +41.4\% \\

        &
        & All layers
        & 15.8842
        & +4.7255
        & +42.3\% \\

        \cmidrule(lr){2-6}

        & Token mean
        & 0--21
        & 11.1832
        & +0.0245
        & +0.2\% \\

        \bottomrule
    \end{tabular}%
    }
\end{table}

\begin{table}[H]
    \centering
    \caption{
    Zero-shot downstream performance under the inference-time interventions.
    All values are percentages; $\Delta$ Avg. is measured in percentage points.
    }
    \label{tab:intervention-downstream}
    \small
    \renewcommand{\arraystretch}{0.8}
    \setlength{\tabcolsep}{4pt}

    \resizebox{0.98\columnwidth}{!}{%
    \begin{tabular}{
        @{\hspace{6pt}}
        lll
        cccccccc
        @{\hspace{6pt}}
    }
        \toprule
        \textbf{Type}
        & \textbf{Action}
        & \textbf{Scope}
        & \textbf{ARC-E}
        & \textbf{ARC-C}
        & \textbf{PIQA}
        & \textbf{HellaSwag}
        & \textbf{MMLU}
        & \textbf{GSM8K}
        & \textbf{Avg.}
        & \textbf{$\Delta$ Avg.} \\
        \midrule

        \multicolumn{3}{@{\hspace{6pt}}l}{Baseline}
        & 87.92
        & 66.38
        & 84.22
        & 85.63
        & 85.44
        & 95.75
        & 84.22
        & --- \\

        \midrule

        \multirow{4}{*}{%
            \makecell[l]{Routing}%
        }
        & \multirow{2}{*}{top-3}
        & $\Hpre$
        & 87.96
        & 66.30
        & 83.90
        & 85.39
        & 85.07
        & 94.39
        & 83.84
        & -0.38 \\

        &
        & $\Hpost$
        & 87.84
        & 66.98
        & 84.22
        & 85.85
        & 85.37
        & 94.77
        & 84.17
        & -0.05 \\

        \cmidrule(lr){2-11}

        &
        \multirow{2}{*}{top-2}
        & $\Hpre$
        & 85.02
        & 62.54
        & 82.81
        & 83.96
        & 81.66
        & 91.81
        & 81.30
        & -2.92 \\

        &
        & $\Hpost$
        & 87.42
        & 66.81
        & 82.97
        & 84.99
        & 85.41
        & 58.15
        & 77.63
        & -6.59 \\

        \midrule

        \multirow{4}{*}{%
            \makecell[l]{Mixer}%
        }
        & \multirow{2}{*}{$\Hres \rightarrow \mI_4$}
        & 22--42
        & 88.05
        & 67.32
        & 84.28
        & 85.81
        & 85.41
        & 94.69
        & 84.26
        & +0.04 \\

        &   
        & All layers
        & 85.56
        & 62.71
        & 83.30
        & 82.98
        & 79.90
        & 91.21
        & 80.94
        & -3.28 \\

        \cmidrule(lr){2-11}

        & Token mean
        & 0--21
        & 87.71
        & 67.49
        & 83.51
        & 85.61
        & 85.14
        & 94.39
        & 83.98
        & -0.25 \\

        \bottomrule
    \end{tabular}%
    }
\end{table}

\paragraph{Routing sparsification.}
Routing exhibits a clear threshold between removing the lowest-weight route for each token and retaining only two routes.
Keeping the three largest realized read or write weights for every token at every site raises perplexity by only $2.4$--$2.7\%$ and changes the average downstream score by at most $0.38$ percentage points.
Keeping only the two largest weights per token, however, raises perplexity by $12.1$--$14.0\%$ and reduces the average downstream score by $2.92$--$6.59$ points; for $\Hpost$, the larger average drop is driven primarily by GSM8K.
The lowest-weight routing path for each token is therefore largely dispensable under the evaluated interventions.
Consistent with this redundancy being local rather than stream-wide, Appendix~\ref{app:lowest-load-route} (Figure~\ref{fig:weakest-routes}) shows that the stream with the lowest mean routing load changes across layers, maps, and sublayer types.

\paragraph{Residual-mixer replacement.}
The functional importance of learned residual mixing closely follows its depth profile.
Replacing the near-identity mixers in layers 22--42 by exact identity maps raises perplexity by only $1.9\%$ and changes the six-task average score by $+0.04$ percentage points.
In contrast, applying the same replacement to layers 0--21 raises perplexity by $41.4\%$, nearly matching the $42.3\%$ increase from replacing the mixers throughout the network.
Fixing each early mixer to its site-specific mean over the C4 diagnostic sequences yields a perplexity of $11.1832$, only $0.2\%$ above the $11.1587$ baseline, and changes the six-task average score from $84.22\%$ to $83.98\%$ ($-0.25$ percentage points).
For this checkpoint and the evaluated metrics, the learned site-specific structure therefore accounts for most of the measured benefit of the early mixers, while their token-wise variation contributes little; the late mixers can instead be replaced by identity with little measured loss, eliminating their Sinkhorn--Knopp iterations at inference time.

Together, these interventions show that training realizes only part of the flexibility afforded by \mhc's four-stream design, with redundancy localized both within routing maps and across depth in the residual mixers.
This is not a wholesale collapse of multi-stream computation: the remaining routing support and the site-specific structure of the early mixers materially affect model quality.

\section{Discussion and Implications}
\label{sec:implications}

Our findings identify two properties of the trained pathway that call for further analysis: routing is concentrated within individual sublayers but reorganizes across depth, and learned residual mixing becomes nearly identity after mid-depth.
We identify candidate optimization mechanisms associated with these patterns and consider implications for future multi-stream designs.

\paragraph{A candidate feedback mechanism for routing concentration.}
Read/write routing and stream representations are learned jointly, creating a possible feedback loop: greater read weight propagates a larger branch-mediated gradient to the parameters producing that stream, while the write map determines which persistent streams receive the branch output.
These interactions may reinforce existing routing preferences, although the realized coefficients remain coupled through a shared token-dependent router.
Appendix~\ref{app:routing-optimization-analysis} formalizes this channel; validating it requires training-time trajectories of routing margins, stream-wise gradients, and representations.

\paragraph{Optimization factors governing residual mixing.}
Movement away from identity depends on both the optimization pressure for cross-stream exchange and the sensitivity of Sinkhorn--Knopp normalization to that pressure.
The final checkpoint cannot distinguish weak functional demand from weak gradient transmission to off-diagonal entries; Appendix~\ref{app:residual-optimization-analysis} formalizes this distinction.
Neither local analysis reconstructs the training dynamics of the checkpoint, whose causal mechanisms require training-time routing and mixer trajectories.

\paragraph{Directions for improving \mhc.}
Recent designs suggest that multi-stream residual parameterization remains unsettled: Hy4-preview adopts four-stream identity Hyper-Connections~\citep{tencent2026hy4}, while Qwen3.8-Flash-Next uses data-dependent Gated Residual across four branches~\citep{qwen2026design}.
These choices accord with our finding that not every degree of freedom in full \mhc is functionally required.
In \mhc, all three maps are predicted from incoming states: this is necessary for \(\Hpre\), but \(\Hpost\) selects where to write the branch output and \(\Hres\) mixes incoming states without conditioning on the computation produced by the branch.
A promising alternative is to retain input-conditioned \(\Hpre\) while allowing \(\Hpost\), and potentially \(\Hres\), to condition additionally on the branch output; evaluating it requires matched training comparisons with full and simplified alternatives.

\section{Conclusion}
\label{sec:conclusion}

We studied how a trained \mhc model realizes the flexibility of its multi-stream residual pathway during inference.
True-forward measurements of DeepSeek-V4-Flash reveal locally concentrated but depth-varying routing, directionally distinct stream representations, and residual exchange that occurs primarily in early layers.
Inference-time interventions further show that the weakest routing weight per token at every site and the late residual mixers can be suppressed with little measured loss, whereas the remaining routing support and the learned structure of the early mixers are functionally important.
Fixing the early mixers to their C4 diagnostic means also nearly preserves model quality, indicating that their site-specific structure contributes more than their token-wise variation on the evaluated metrics.

The four-stream pathway thus exhibits structured under-utilization rather than uniform participation or complete stream collapse.
More broadly, architectural width does not determine effective multi-stream capacity; its realization depends on the component- and depth-specific organization learned during training.
Our analysis identifies candidate mechanisms whose causal roles require training-time routing and mixer trajectories.

\newpage
\bibliography{iclr2026_conference}

@inproceedings{he2016resnet,
  title={Deep Residual Learning for Image Recognition},
  author={He, Kaiming and Zhang, Xiangyu and Ren, Shaoqing and Sun, Jian},
  booktitle={CVPR},
  year={2016}
}

@article{raffel2020t5,
  title={Exploring the Limits of Transfer Learning with a Unified Text-to-Text Transformer},
  author={Raffel, Colin and Shazeer, Noam and Roberts, Adam and Lee, Katherine and Narang, Sharan and Matena, Michael and Zhou, Yanqi and Li, Wei and Liu, Peter J},
  journal={JMLR},
  volume={21},
  number={140},
  pages={1--67},
  year={2020}
}

@article{zhu2024hyperconnections,
  title={{Hyper-Connections}},
  author={Zhu, Defa and Huang, Hongzhi and Huang, Zihao and Zeng, Yutao and Mao, Yunyao and Wu, Banggu and Min, Qiyang and Zhou, Xun},
  journal={arXiv preprint arXiv:2409.19606},
  year={2024},
  eprint={2409.19606},
  archivePrefix={arXiv},
  primaryClass={cs.LG},
  url={https://arxiv.org/abs/2409.19606}
}

@article{xie2026mhc,
  title={{mHC}: {Manifold-Constrained Hyper-Connections}},
  author={Xie, Zhenda and Wei, Yixuan and Cao, Huanqi and Zhao, Chenggang and Deng, Chengqi and Li, Jiashi and Dai, Damai and Gao, Huazuo and Chang, Jiang and Yu, Kuai and Zhao, Liang and Zhou, Shangyan and Xu, Zhean and Zhang, Zhengyan and Zeng, Wangding and Hu, Shengding and Wang, Yuqing and Yuan, Jingyang and Wang, Lean and Liang, Wenfeng},
  journal={arXiv preprint arXiv:2512.24880},
  year={2025},
  eprint={2512.24880},
  archivePrefix={arXiv},
  primaryClass={cs.CL},
  url={https://arxiv.org/abs/2512.24880}
}

@article{sinkhorn1967concerning,
  title={Concerning nonnegative matrices and doubly stochastic matrices},
  author={Sinkhorn, Richard and Knopp, Paul},
  journal={Pacific Journal of Mathematics},
  volume={21},
  number={2},
  pages={343--348},
  year={1967}
}

@misc{alimaskina2026streamcollapse,
      title={Analyzing Stream Collapse in {Hyper-Connections}: From Diagnosis to Mitigation}, 
      author={Ekaterina Alimaskina and Gleb Molodtsov and Aleksandr Beznosikov},
      year={2026},
      eprint={2606.03483},
      archivePrefix={arXiv},
      primaryClass={cs.LG},
      url={https://arxiv.org/abs/2606.03483}, 
}

@misc{karpathy2022nanogpt,
      author={Andrej Karpathy},
      title={{nanoGPT}: The Simplest, Fastest Repository for Training and Fine-Tuning Medium-Sized {GPTs}},
      year={2022},
      howpublished={\url{https://github.com/karpathy/nanoGPT}},
      note={GitHub repository},
}

@misc{deepseekai2026deepseekv4,
      title={{DeepSeek-V4}: Towards Highly Efficient {Million-Token} Context Intelligence},
      author={{DeepSeek-AI}},
      year={2026},
      eprint={2606.19348},
      archivePrefix={arXiv},
      primaryClass={cs.CL},
      url={https://arxiv.org/abs/2606.19348},
}

@article{clark2018think,
  title={Think You Have Solved Question Answering? {Try} {ARC}, the {AI2} Reasoning Challenge},
  author={Clark, Peter and Cowhey, Isaac and Etzioni, Oren and Khot, Tushar and Sabharwal, Ashish and Schoenick, Carissa and Tafjord, {{\O}}yvind},
  journal={arXiv preprint arXiv:1803.05457},
  year={2018},
  url={https://arxiv.org/abs/1803.05457}
}

@inproceedings{bisk2020piqa,
  title={{PIQA}: Reasoning about Physical Commonsense in Natural Language},
  author={Bisk, Yonatan and Zellers, Rowan and Le Bras, Ronan and Gao, Jianfeng and Choi, Yejin},
  booktitle={AAAI},
  year={2020},
  url={https://arxiv.org/abs/1911.11641}
}

@inproceedings{zellers2019hellaswag,
  title={{HellaSwag}: Can a Machine Really Finish Your Sentence?},
  author={Zellers, Rowan and Holtzman, Ari and Bisk, Yonatan and Farhadi, Ali and Choi, Yejin},
  booktitle={Proceedings of the 57th Annual Meeting of the Association for Computational Linguistics},
  pages={4791--4800},
  year={2019},
  doi={10.18653/v1/P19-1472},
  url={https://aclanthology.org/P19-1472/}
}

@inproceedings{hendrycks2021measuring,
  title={Measuring Massive Multitask Language Understanding},
  author={Hendrycks, Dan and Burns, Collin and Basart, Steven and Zou, Andy and Mazeika, Mantas and Song, Dawn and Steinhardt, Jacob},
  booktitle={ICLR},
  year={2021},
  url={https://arxiv.org/abs/2009.03300}
}

@article{cobbe2021training,
  title={Training Verifiers to Solve Math Word Problems},
  author={Cobbe, Karl and Kosaraju, Vineet and Bavarian, Mohammad and Chen, Mark and Jun, Heewoo and Kaiser, Lukasz and Plappert, Matthias and Tworek, Jerry and Hilton, Jacob and Nakano, Reiichiro and Hesse, Christopher and Schulman, John},
  journal={arXiv preprint arXiv:2110.14168},
  year={2021},
  url={https://arxiv.org/abs/2110.14168}
}

@inproceedings{frantar2023gptq,
  title={{GPTQ}: Accurate Post-Training Quantization for Generative Pre-trained Transformers},
  author={Frantar, Elias and Ashkboos, Saleh and Hoefler, Torsten and Alistarh, Dan},
  booktitle={International Conference on Learning Representations},
  year={2023},
  url={https://arxiv.org/abs/2210.17323}
}

@inproceedings{paszke2019pytorch,
  title={{PyTorch}: An Imperative Style, High-Performance Deep Learning Library},
  author={Paszke, Adam and Gross, Sam and Massa, Francisco and Lerer, Adam and Bradbury, James and Chanan, Gregory and Killeen, Trevor and Lin, Zeming and Gimelshein, Natalia and Antiga, Luca and Desmaison, Alban and K{\"o}pf, Andreas and Yang, Edward and DeVito, Zachary and Raison, Martin and Tejani, Alykhan and Chilamkurthy, Sasank and Steiner, Benoit and Fang, Lu and Bai, Junjie and Chintala, Soumith},
  booktitle={Advances in Neural Information Processing Systems},
  volume={32},
  year={2019}
}

@inproceedings{wolf2020transformers,
  title={Transformers: State-of-the-Art Natural Language Processing},
  author={Wolf, Thomas and Debut, Lysandre and Sanh, Victor and Chaumond, Julien and Delangue, Clement and Moi, Anthony and Cistac, Perric and Ma, Clara and Jernite, Yacine and Plu, Julien and Xu, Canwen and Le Scao, Teven and Gugger, Sylvain and Drame, Mariama and Lhoest, Quentin and Rush, Alexander M.},
  booktitle={Proceedings of the 2020 Conference on Empirical Methods in Natural Language Processing: System Demonstrations},
  pages={38--45},
  year={2020}
}

@misc{gao2023lmeval,
  title={A Framework for Few-Shot Language Model Evaluation},
  author={Gao, Leo and Tow, Jonathan and Abbasi, Baber and Biderman, Stella and Black, Sid and DiPofi, Anthony and Foster, Charles and Golding, Laurence and Hsu, Jeffrey and Le Noac'h, Alain and Li, Haonan and McDonell, Kyle and Muennighoff, Niklas and Ociepa, Chris and Phang, Jason and Reynolds, Laria and Schoelkopf, Hailey and Skowron, Aviya and Sutawika, Lintang and Tang, Eric and Thite, Anish and Wang, Ben and Wang, Kevin and Zou, Andy},
  publisher={Zenodo},
  year={2023},
  version={v0.4.0},
  doi={10.5281/zenodo.10256836},
  url={https://zenodo.org/records/10256836}
}

@misc{tencent2026hy4,
  title={{Hy4-preview} Model Card},
  author={{Tencent Hy Team}},
  year={2026},
  howpublished={\url{https://huggingface.co/tencent/Hy4-preview}},
  note={Accessed September 2026}
}

@techreport{qwen2026design,
  title={On the Design of {Qwen3.8-Next} Architecture: Evaluation, Efficiency, and Training Stability},
  author={{Qwen Team}},
  institution={Alibaba Group},
  year={2026},
  month={August},
  eprint={2608.30320},
  archivePrefix={arXiv},
  primaryClass={cs.CL},
  url={https://arxiv.org/abs/2608.30320}
}
\bibliographystyle{iclr2026_conference}

\clearpage
\appendix
\section{Additional Diagnostic Results}
\label{app:additional-diagnostics}

\subsection{Diagnostics by Depth and Sublayer Type}
\label{app:diagnostics-depth-site}

Table~\ref{tab:diagnostics-depth-site} places the routing, representation, and residual-mixing diagnostics under a common depth split.
The reduction in residual mixing is consistent across attention and FFN sites, while the change in routing breadth is concentrated in the write maps, particularly at FFN sites.

\begin{table}[H]
    \centering
    \caption{
    Mean diagnostics by depth range and sublayer type.
    Winner consistency (WC) and effective stream count ($n_{\mathrm{eff}}$) are averaged over the indicated routing sites; cosine similarity is additionally averaged over the six stream pairs.
    }
    \label{tab:diagnostics-depth-site}
    \small
    \renewcommand{\arraystretch}{1.08}
    \setlength{\tabcolsep}{9pt}
    \resizebox{0.94\linewidth}{!}{%
    \begin{tabular}{llcccccc}
        \toprule
        & & \multicolumn{2}{c}{\textbf{Winner consistency}}
        & \multicolumn{2}{c}{\textbf{$n_{\mathrm{eff}}$}}
        & \textbf{Cosine}
        & \textbf{$\delta(\Hres)$} \\
        \cmidrule(lr){3-4}\cmidrule(lr){5-6}
        \textbf{Layers} & \textbf{Site}
        & $\Hpre$ & $\Hpost$
        & $\Hpre$ & $\Hpost$
        & & \\
        \midrule
        \multirow{2}{*}{0--21}
        & Attention & 0.874 & 0.970 & 2.174 & 1.317 & 0.456 & 0.045 \\
        & FFN       & 0.905 & 0.909 & 1.768 & 1.618 & 0.438 & 0.047 \\
        \midrule
        \multirow{2}{*}{22--42}
        & Attention & 0.833 & 0.917 & 2.104 & 1.836 & 0.385 & 0.010 \\
        & FFN       & 0.872 & 0.823 & 1.947 & 2.360 & 0.378 & 0.008 \\
        \bottomrule
    \end{tabular}%
    }
\end{table}

\subsection{Identity of the Lowest-Load Route}
\label{app:lowest-load-route}

\begin{figure}[H]
    \centering
    \includegraphics[width=0.95\linewidth]{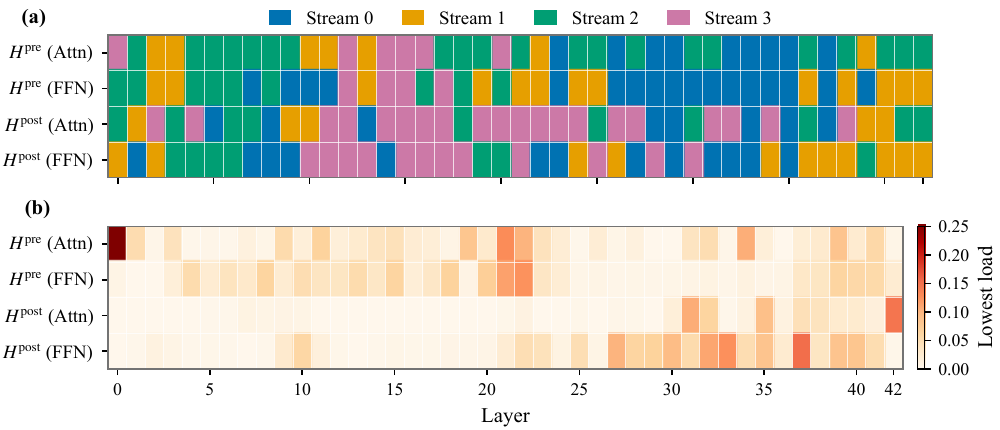}
    \caption{
    Lowest-load routing stream across depth.
    (a) Identity of the stream with the smallest mean normalized routing load at each layer, map, and sublayer type.
    (b) Its corresponding mean load.
    The weakest stream changes across sites rather than remaining globally fixed.
    }
    \label{fig:weakest-routes}
\end{figure}

Figure~\ref{fig:weakest-routes} ranks streams by their mean normalized routing load at each site.
All four streams appear as the lowest-load stream somewhere in each map/site combination, so the locally weak route does not correspond to a single stream that can be removed throughout the network.
Its mean load is generally small but varies across depth, favoring site-adaptive sparsification over the removal of a globally fixed stream.

\clearpage
\section{Layerwise Residual Mixers}
\label{app:layerwise-hres}

\begin{figure}[H]
    \centering
    \includegraphics[width=0.92\linewidth]{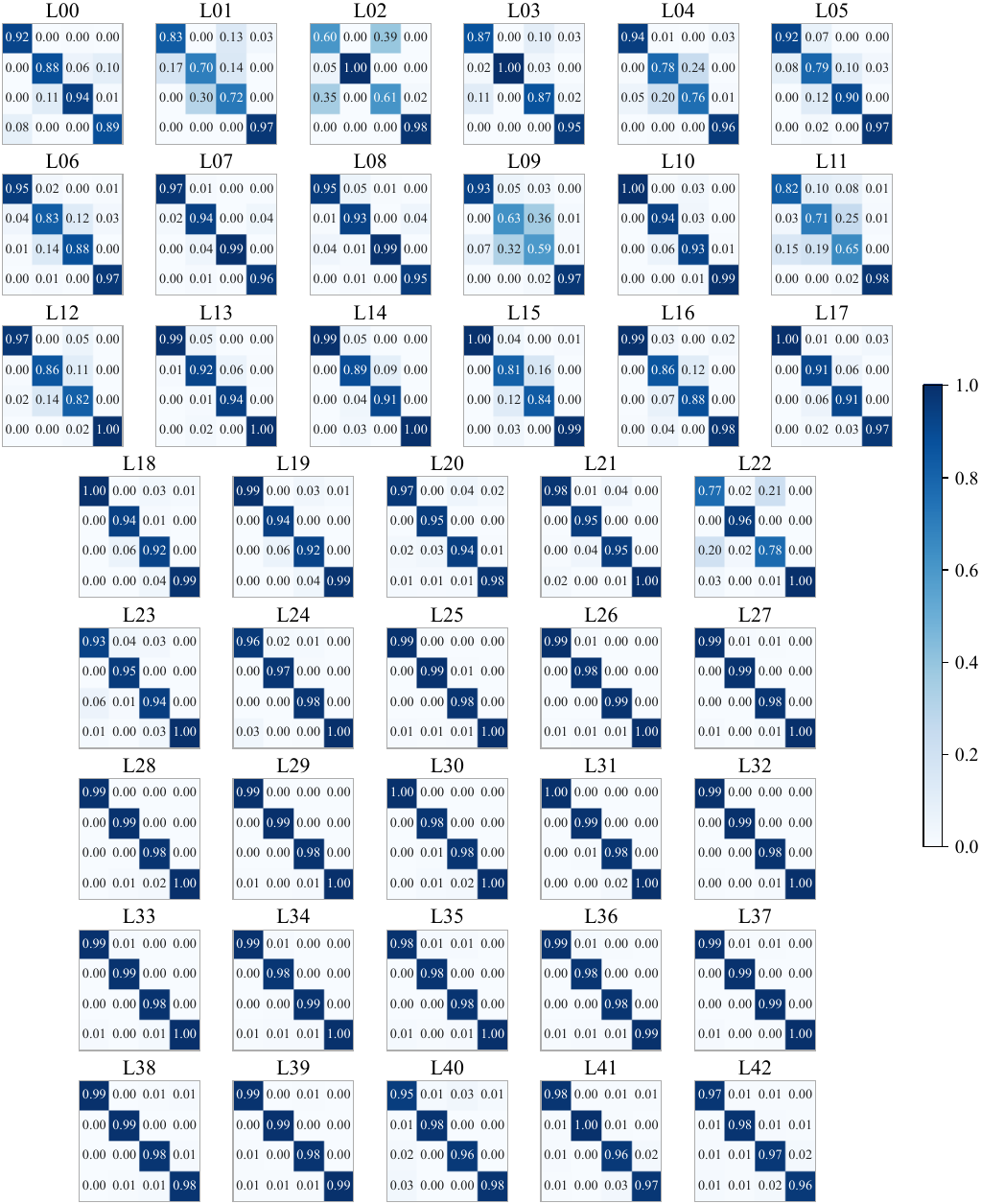}
    \caption{Token-averaged realized residual mixers at the attention site of every layer.}
    \label{fig:hres-attn-all}
\end{figure}

\begin{figure}[H]
    \centering
    \includegraphics[width=0.92\linewidth]{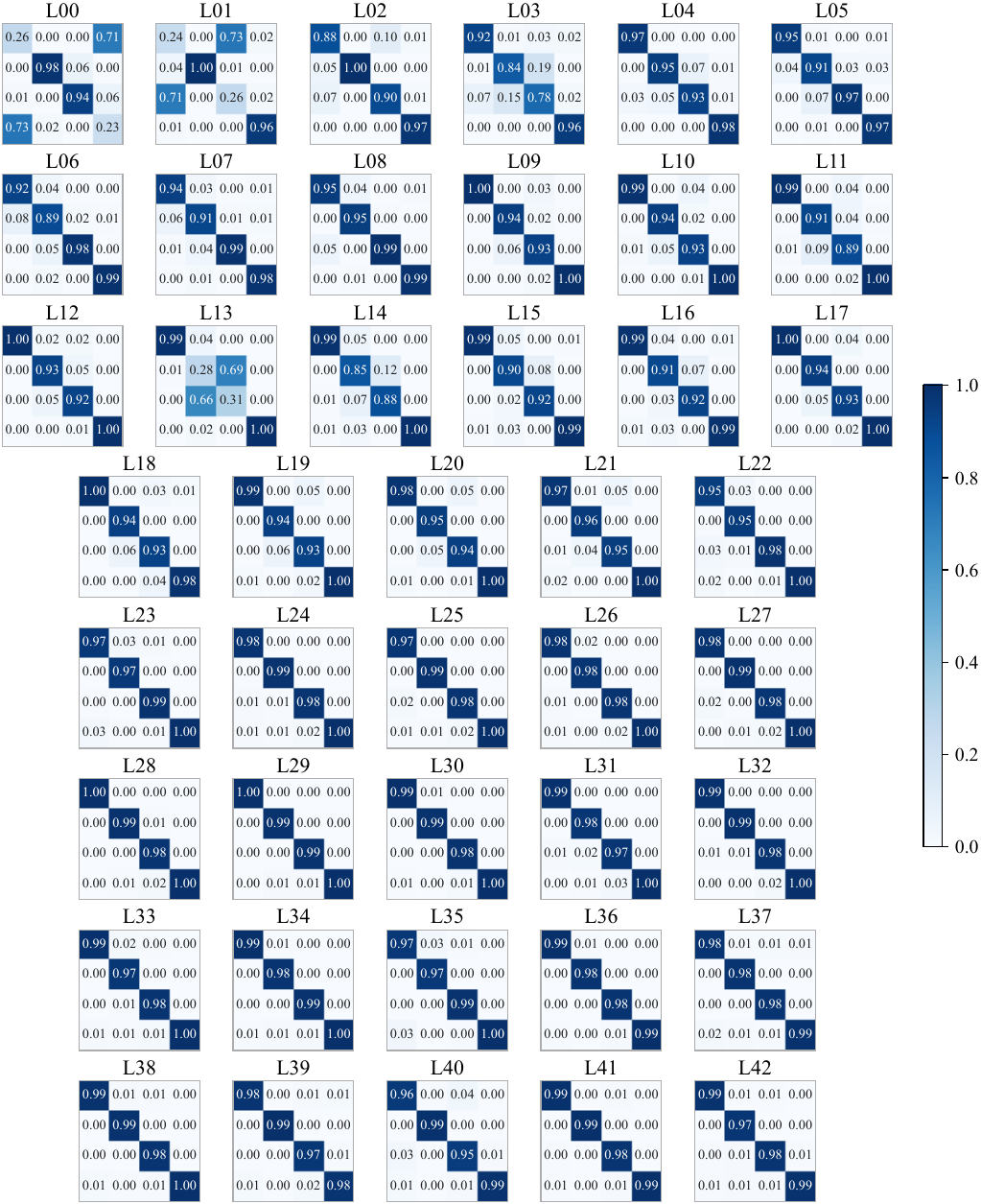}
    \caption{Token-averaged realized residual mixers at the FFN site of every layer.}
    \label{fig:hres-ffn-all}
\end{figure}

Figures~\ref{fig:hres-attn-all} and~\ref{fig:hres-ffn-all} show the token-averaged realized residual mixer at every layer, separately for attention and FFN sites.
They complement the depth profile and range averages in Figure~\ref{fig:hres-structure} by exposing which stream pairs account for the residual exchange at each site.

\paragraph{Attention sites.}
Figure~\ref{fig:hres-attn-all} shows that the off-diagonal structure is both depth- and pair-specific.
The most visible exchange occurs at several early and middle sites, including layers 1--2, 9, and 11, followed by a final pronounced deviation at layer 22.
Beyond this point, the matrices are consistently near diagonal, with only small site-level fluctuations through layer 42.

\clearpage
\paragraph{FFN sites.}
Figure~\ref{fig:hres-ffn-all} exhibits the same broad reduction in residual exchange but a different set of high-mixing sites.
The strongest off-diagonal weights occur at layers 0--1 and 13, whereas the mixers from layer 23 onward remain close to identity.
The locations and stream pairs responsible for early mixing therefore differ between attention and FFN, while the late near-identity regime is shared by both sublayer types.

\clearpage
\section{Local Optimization Analysis}
\label{app:optimization-analysis}

\subsection{Routing Concentration}
\label{app:routing-optimization-analysis}

For a single token and sublayer, write the realized read and write maps as $\vr=(r_1,\ldots,r_n)$ and $\vw=(w_1,\ldots,w_n)$, such that
\(\vz=\sum_j r_j\vx^{(j)}\), \(\vy=F(\vz)\), and the branch adds \(w_i\vy\) to output stream $i$.
Let \(\vg_z=\partial\mathcal{L}/\partial\vz\) and \(\vg_i^+=\partial\mathcal{L}/\partial\vx^{+(i)}\).
Conditioned on the realized routing coefficients, the direct branch-mediated gradients are
\begin{equation}
\label{equ:routing-gradients}
    \left.\frac{\partial\mathcal{L}}{\partial\vx^{(j)}}\right|_{\mathrm{branch}}
    = r_j\vg_z,
    \qquad
    \frac{\partial\mathcal{L}}{\partial r_j}
    = \left\langle\vg_z,\vx^{(j)}\right\rangle,
    \qquad
    \frac{\partial\mathcal{L}}{\partial w_i}
    = \left\langle\vg_i^+,\vy\right\rangle.
\end{equation}
At the level of the realized read coefficients, define the negative-gradient direction
\begin{equation}
\label{equ:read-descent-score}
    d_{r_j}
    \coloneqq
    -\frac{\partial\mathcal{L}}{\partial r_j}
    = -\left\langle\vg_z,\vx^{(j)}\right\rangle.
\end{equation}
Stream $j$ is favored over stream $k$ by this local direction when $d_{r_j}>d_{r_k}$.
A larger $r_j$ also scales the branch-mediated gradient propagated through $\vx^{(j)}$, and hence the gradient received by the upstream parameters that produce this state.
If the resulting parameter update makes $d_{r_j}$ more favorable on subsequent examples, the existing read preference can be reinforced.
The write map participates in the same feedback by controlling which persistent states receive the branch output and influence later computation.
Because the realized routing coefficients are coupled through a shared token-dependent router, these descent scores identify a possible feedback channel rather than the router parameters' actual update.

\subsection{Residual Mixing}
\label{app:residual-optimization-analysis}

Write the realized mixer at a token and sublayer as $\mH^{\mathrm{res}}=\mathcal{S}(\mA)$, where $\mA$ contains the pre-Sinkhorn logits and $\mathcal{S}$ denotes Sinkhorn normalization, and let
\begin{equation}
\label{equ:sinkhorn-jacobian}
    \vh \coloneqq \operatorname{vec}(\mH^{\mathrm{res}}),
    \qquad
    \va \coloneqq \operatorname{vec}(\mA),
    \qquad
    \mJ_{\mathcal{S}} \coloneqq
    \frac{\partial \vh}{\partial \va},
    \qquad
    \vg_h \coloneqq \frac{\partial\mathcal{L}}{\partial\vh}.
\end{equation}
For an idealized local gradient step on the logits, first-order expansion gives
\begin{equation}
\label{equ:sinkhorn-local-update}
    \Delta\vh
    \approx
    -\eta\mJ_{\mathcal{S}}\mJ_{\mathcal{S}}^\top\vg_h.
\end{equation}
If $\mP_{\mathrm{off}}$ projects a vectorized mixer onto its off-diagonal entries, then
\begin{equation}
\label{equ:sinkhorn-offdiag-bound}
    \left\|\mP_{\mathrm{off}}\Delta\vh\right\|_2
    \le
    \eta
    \left\|\mP_{\mathrm{off}}\mJ_{\mathcal{S}}\right\|_2
    \left\|\mJ_{\mathcal{S}}\right\|_2
    \left\|\vg_h\right\|_2.
\end{equation}
Movement away from identity can therefore remain small either because the loss gradient supplies little pressure for cross-stream residual exchange or because the local Sinkhorn Jacobian weakly transmits that pressure to off-diagonal entries.
The final checkpoint does not reveal which factor governed the late-layer mixers or how they reached the observed regime.
\end{document}